\documentclass{article} 
\usepackage{iclr2020_conference,times}

\PassOptionsToPackage{numbers, sort&compress}{natbib}

\usepackage[utf8]{inputenc} 
\usepackage[T1]{fontenc}    
\usepackage{xcolor}         
\definecolor{custom-blue}{RGB}{6,69,173} 
\usepackage[colorlinks, allcolors=custom-blue]{hyperref}
\usepackage{hyperref}       
\usepackage{url}            
\usepackage{booktabs}       
\usepackage{amsfonts}       
\usepackage{nicefrac}       
\usepackage{microtype}      

\usepackage{multirow, makecell}
\usepackage{graphicx}
\usepackage{wrapfig}
\graphicspath{{./}}
\usepackage{subcaption}
\usepackage{adjustbox}

\usepackage{footnote}
\makesavenoteenv{tabular}
\makesavenoteenv{table}

\usepackage{array}
\newcolumntype{L}[1]{>{\raggedright\let\newline\\\arraybackslash\hspace{0pt}}m{#1}}
\newcolumntype{C}[1]{>{\centering\let\newline\\\arraybackslash\hspace{0pt}}m{#1}}
\newcolumntype{R}[1]{>{\raggedleft\let\newline\\\arraybackslash\hspace{0pt}}m{#1}}

\usepackage{siunitx}
\usepackage{mathtools}
\usepackage{autonum}
\usepackage{amsthm}
\usepackage{enumitem}
\usepackage{algorithm}
\usepackage{algorithmic}
\usepackage{eqparbox}

\usepackage{setspace}

\DeclareMathOperator*{\E}{\mathbb{E}}
\DeclareMathOperator{\relu}{ReLU}
\newcommand{\loss}{{\ell}}
\DeclarePairedDelimiterX{\inner}[2]{\langle}{\rangle}{#1, #2} 

\newcommand{\norm}[1]{\Vert#1\Vert}

\DeclareMathOperator{\grad}{\nabla}
\DeclareMathOperator{\sgn}{sign}

\theoremstyle{Proposition}
\newtheorem{theorem}{Theorem}[section]

\newtheorem{proposition}[theorem]{Proposition}

\theoremstyle{definition}

\theoremstyle{remark}
\newtheorem{remark}[theorem]{Remark}

\title{Scaleable input gradient regularization for adversarial robustness}

\author{%
  Chris Finlay\\
  McGill University \\
  \texttt{christopher.finlay@mail.mcgill.ca} \\
  \And
  Adam M Oberman \\
  McGill University \\
  \texttt{adam.oberman@mcgill.ca} \\
}

\iclrfinalcopy 

\begin{document}

\maketitle

\begin{abstract}
In this work we revisit gradient regularization for adversarial robustness with some new ingredients.  First, we derive new per-image theoretical robustness bounds based on local gradient information. These bounds strongly motivate input gradient regularization.  Second, we implement a scaleable version of input gradient regularization which avoids double backpropagation: adversarially robust ImageNet models are trained in 33 hours on four consumer grade GPUs.  Finally, we show experimentally and through theoretical certification that input gradient regularization is competitive with adversarial training. Moreover we demonstrate that gradient regularization does not lead to gradient obfuscation or gradient masking.
\end{abstract}

\section{Introduction}
Neural networks are vulnerable to \emph{adversarial attacks}. These are small
(imperceptible to the human eye) perturbations of an image which cause a network
to misclassify  the image \citep{biggio2013, szegedy2013intriguing,
goodfellow2014explaining}.   The threat posed by adversarial attacks
must be addressed before these methods can be deployed in error-sensitive and
security-based applications \citep{potember2017perspectives}.

Building adversarially robust models is an optimization problem with two
objectives: (i) maintain test accuracy on clean unperturbed images, and (ii)
be robust to large adversarial perturbations.
The present state-of-the-art method for adversarial defence, adversarial
training \citep{szegedy2013intriguing,
goodfellow2014explaining, tramer_ensemble_2017, madry_2017,miyato2018virtual},
in which models are trained on perturbed images, 
offers robustness at the expense of test accuracy \citep{tsipras2018robustness}. 
Up until the recent method of \citet{shafahi2019}, multi-step adversarial training
had taken many days to train on large datasets
\citep{kannan2018adversarial,xie2018}.

Assessing the \emph{empirical} effectiveness of an adversarial defence requires careful
testing with multiple attacks \citep{goodfellow_2018}.  Furthermore, existing
defences are vulnerable to new, stronger attacks:
\citet{carlini2017adversarial} and \citet{athalye2018obfuscated} advocate
designing specialized attacks to circumvent prior defences, while \citet{uesato2018}
warn against using weak attacks to evaluate robustness.
This has led the community to develop \emph{theoretical} tools to certify
adversarial robustness. Several certification approaches have been proposed:
through linear programming \citep{wong2018a,wong2018b} or mixed-integer linear-programming
\citep{xiao2018training}; semi-definite relaxation
\citep{raghunathan2018, raghunathan2018b};  randomized smoothing
\citep{li2018second,cohen2019}; or estimates of the local
Lipschitz constant
\citep{hein2017formal,weng2018evaluating,tsuzuku_margin_2018}. The latter two
approaches have scaled well to ImageNet-1k.  Notably, randomized smoothing
with adversarial training
has shown great promise for rigorous certification against attacks in the $\ell_2$ norm
\citep{salman2019}, although it is not yet clear how randomized smoothing may be
adapted to other norms.

In practice, certifiably robust networks often perform worse than
adversarially trained models, which in general lack theoretical guarantees.
In this article, we work towards bridging the gap between theoretically robust
networks and
empirically effective training methods. Our approach relies on minimizing a loss regularized against
large input gradients
\begin{equation}\label{modified_loss}
  \E_{(x,y)\sim \mathbb P} \left[\mathcal L(f(x;w),y) + \frac{\lambda}{2} \norm{\grad_x \mathcal L(f(x;w),y)}^2_* \right]
\end{equation}
where $\norm{\cdot}_*$ is dual to the one   measuring adversarial
attacks (for example the $\ell_1$ norm for attacks measured in the
$\ell_\infty$ norm). 
Heuristically, making loss gradients small should make gradient based attacks
more challenging.

\citet{drucker1991double} implemented gradient
regularization using `double backpropagation', which has been shown to improve
model generalization \citep{novak2018}. 
It has been used to improve the stability of GANs
\citep{roth2017gan,nagarajan2017} and to promote learning robust features with contractive auto-encoders
\citep{contractive-ae}.
While it has been proposed for  adversarial attacks robustness
\citep{lyu15,ross2018,roth2018,ororbia17,hein2017formal,jakubovitz2018,gabriel2018},
experimental evidence has been mixed; in particular, input gradient regularization has so far not been competitive with multi-step adversarial training. 

In particular, the heuristic that ``gradients should be small'' has motivated
several other approaches to adversarial robustness which were later shown to suffer from problems of
\emph{gradient obfuscation} \citep{athalye2018obfuscated}. This is the phenomenon
whereby gradients are in some sense hidden or minimized, so that gradient-based
attacks fail to produce adversarial examples.  Due to this phenomenon, the 
community now greets adversarial robustness methods based on notions of small
gradients with a healthy dose of skepticism. It is therefore necessary to
demonstrate new methods do not suffer from gradient obfuscation
\citep{carlini2019}, both empirically
and theoretically: empirically, by attacking models with gradient-free methods
or non-local attacks;
and theoretically using tight certification techniques.

Our main contributions in this work are the following. First, we motivate using input gradient
regularization \emph{of the loss} by deriving new theoretical robustness bounds.
These bounds use the gradient of the loss and an estimate of the error of the linear
approximation of the loss called the \emph{modulus of continuity}. These
theoretical bounds show that small loss gradient and a small modulus of
continuity are sufficient conditions for adversarial  robustness. 
Second, we empirically show that input gradient
regularization is competitive with standard adversarial training
\citep{madry_2017}, at a fraction of the training time. 
We verify that training with gradient regularization does not lead to gradient
obfuscation, by attacking our trained models with gradient-free methods, and
through certification bounds.
Finally, we scale input gradient
regularization
to ImageNet-1k \citep{imagenet} by using finite differences to estimate the gradient
regularization term, rather than double backpropagation (which does not scale).
This allows us to train adversarially robust networks on ImageNet-1k with only
50\% more training time than that of standard (non robust) training.

\section{Adversarial robustness bounds from the loss}

\subsection{Background}

Much effort has been directed towards determining theoretical lower bounds on
the minimum sized perturbation necessary to perturb an image so that it is
misclassified by a model. One promising approach, proposed by
\citet{hein2017formal} and \citet{weng2018evaluating}, and which has scaled well to
ImageNet-1k, is to use the Lipschitz constant of the model. In this section, we
build upon these ideas: we propose using the Lipschitz constant of a suitable
loss, designed to measure classification errors.  
Additionally, we propose a tighter lower bound which uses the \emph{local} gradient
information of the loss, and an error bound on the first-order
approximation of the loss.


 Our notation is as follows.  Write $y = f(x;w)$ for a model which takes input
 vectors $x$ to label probabilities, with parameters $w$.  Let $\mathcal L(y_1,y_2)$
 be the loss and write $\loss(x) := \mathcal L(f(x,w),y)$, for the loss of a
 model $f$. 
  
 Finding an adversarial perturbation is interpreted as a global minimization problem:
 find the closest image to a clean image, in some specified norm, that is
 misclassified by the model
 \begin{eqnarray}\label{eq:global}
   \min_v \norm{v} & \text{subject to } f(x+v) \text{ misclassified}
 \end{eqnarray}
 However, \eqref{eq:global} is a difficult and costly non-smooth,
 non-convex optimization
 problem. Instead, \citet{goodfellow2014explaining} proposed solving a surrogate
 problem: find a perturbation $v$ of a clean image $x$ that maximizes the loss, subject to the condition
 that the perturbation be inside a norm-ball of radius $\delta$ around the clean
 image. The surrogate problem is written
 \begin{eqnarray}\label{eq:ideal_attack}
   \max_v \loss(x+v) - c(v); & \text{where } c(v) = \begin{cases} 0 & \text{if } \norm{v}\leq
     \delta \\ \infty &\text{otherwise} \end{cases}
\end{eqnarray}
The hard constraint $c(v)$ forces perturbations to be inside the
norm-ball centred at the clean image $x$. 
Ideally, solutions of this surrogate problem \eqref{eq:ideal_attack} will closely align with solutions of
the original more difficult global minimization problem.   However, the hard
constraint in \eqref{eq:ideal_attack} forces a particular scale: it may miss
attacks which would succeed with only a slightly bigger norm.
Additionally, the maximization problem \eqref{eq:ideal_attack} does not force
misclassification; it only asks that the loss be increased. 
%

The advantage of \eqref{eq:ideal_attack} is that it may be solved with
gradient-based methods: present best-practice is to use variants of projected
gradient descent (PGD), such as the iterative fast-signed gradient method
\citep{kurakin2016adversarial,madry_2017} when attacks are measured in the
$\ell_\infty$ norm.  However, gradient-based methods are not always effective:
on non-smooth networks, such as those built of $\relu$ activation functions, a
small gradient does not guarantee that the loss remains small locally.  That is, $\relu$ networks may increase rapidly with a very
small perturbation, even when local gradients are small. This
deficiency was identified in \citep{papernot_limitations_2015} and expanded on
by \citet{athalye2018obfuscated}.
In this scenario, PGD methods will fail
to locate these worst-case perturbations and gives
a false impression of robustness. \citet{carlini_towards_2016} avoid this
scenario by incorporating decision boundary information into the loss; others
solve \eqref{eq:global} directly
\citep{brendel2018decisionbased,boundaryplus,logbarrier} using gradient-free
methods or non-local attacks.

\subsection{Derivation of theoretical lower bounds}\label{sec:bounds}
This leads us to consider the following compromise between \eqref{eq:global} and
\eqref{eq:ideal_attack}.  Consider the  following modification of the \citet{carlini_towards_2016}   loss 
$\loss(x) = \max_{i\neq c} f_{i}(x) - f_c(x)$, where $c$ is the index of the
correct label, and
$f_{i}(x)$ is the model output for the $i$-th label.
This loss has the appealing property that the sign of the loss determines if the classification is correct.  
Adversarial attacks are found by minimizing
\begin{eqnarray}\label{eq:min_attack}
  \min_v \norm{v} &\text{subject to } \loss(x+v) \geq \loss_0
\end{eqnarray}
The constant $\loss_0$ determines when classification is incorrect; for the
modified Carlini-Wagner loss, $\ell_0=0$. Problem \eqref{eq:min_attack} is closer to the true
problem~\eqref{eq:global} and will always find an adversarial image.
We use \eqref{eq:min_attack} to derive theoretical lower bounds on
the minimum size perturbation necessary to misclassify an image.
Suppose the loss is $L$-Lipschitz with respect to model input. Then we have the estimate
\begin{equation}\label{eq:Lip}
  \loss(x+v) \leq  \loss(x) + L \norm{v} 
\end{equation}
Now suppose $v$ \emph{is} adversarial, with minimum adversarial loss $\loss(x+v) = \loss_0$. Then
rearranging \eqref{eq:Lip}, we obtain the lower bound $\norm{v} \geq \frac{1}{L}\left(\loss_0
- \loss(x)\right).$

Unfortunately, the Lipschitz constant is a global quantity, and ignores
local gradient information; see for example \citet{huster2018}. Thus this bound
can be quite poor, even when networks have a small Lipschitz constant.
A tighter theoretical bound may instead be achieved by using (i) the local
gradient, and (ii) a bound on the error of a first order approximation of the
loss.
At image $x$ and perturbations up to size $\varepsilon$, the error of a first
order approximation of the loss  is bounded above by the modulus of continuity
\cite[Ch 3]{timan_theory_1994}
\begin{equation}\label{eq:mod}
  \omega(\varepsilon) = \sup_{x,\,\norm{v}\leq \varepsilon} \loss(x+v) - \left[\loss(x)
  + \inner{v}{\nabla \loss(x)} \right]
\end{equation}
Intuitively, this measures the maximum amount the gradient can change over
perturbations of size $\varepsilon$. Note that the quantity is defined for any
norm. 
Using this quantity, for perturbations of size $\varepsilon$ we have the estimate
\begin{align}
  \loss(x+v) &\leq  \loss(x) + \inner{v}{\nabla \loss(x)} + \omega(\varepsilon) \\
             &\leq  \loss(x) + \norm{v}\norm{\nabla \loss(x)}_* + \omega(\varepsilon) \label{eq:modulbound}
\end{align}
As before, assume $v$ is adversarial, with minimum adversarial loss
$\loss(x+v) = \loss_0$. Rearranging \eqref{eq:modulbound}, we obtain the lower
bound $\norm{v} \geq \frac{1}{\norm{\nabla \loss(x)}_*}\left(\loss_0 - \loss(x) -
\omega(\varepsilon) \right)$.

\begin{remark}
  If the model and the loss are twice continuously differentiable, the modulus
  of continuity can be derived from the maximum curvature. However
  $\omega(\varepsilon)$ is defined even when the model and the loss are not
  twice differentiable.
\end{remark}

We have proved the following.
\begin{proposition}\label{prop:cert}
  Suppose the loss $\loss(x)$ is Lipschitz continuous with respect to model
  input $x$, with Lipschitz constant
  $L$.  Let $\loss_0$ be such that if $\loss(x) < \loss_0$, the model is always
  correct. Then a lower bound on the minimum magnitude of perturbation $v$ necessary to
  adversarially perturb an image $x$ is   
  \begin{equation}
    \tag{$L$-bound}
    \norm{v}  \geq \frac{\max\{\loss_0 - \loss(x), 0\}}{L}
\label{eq:bound1}
  \end{equation}
  Suppose in addition the modulus of continuity $\omega(\varepsilon)$ is
  defined as in \eqref{eq:mod}. Then the minimum adversarial distance is bounded
  below by $\varepsilon$ provided the following inequality holds
\begin{equation}\label{eq:bound2}
  \tag{$\omega$-bound}
  \frac{\loss_0 - \loss(x) -
  \omega(\varepsilon) }{\norm{\nabla \loss(x)}_*} \geq \varepsilon
\end{equation}
\end{proposition}


Proposition \ref{prop:cert} motivates the need for input gradient regularization. 
The Lipschitz constant $L$ is the maximum gradient norm of \emph{the loss} over all
inputs. Therefore \eqref{eq:bound1} says that a regularization term encouraging
small gradients (and so reducing $L$) should increase the minimum adversarial
distance. This aligns with \citep{hein2017formal}, who proposed the
cross-Lipschitz regularizer, penalizing
networks with large Jacobians in order to shrink the Lipschitz constant of \emph{the
network}.

However, this is not enough: the gap $\loss_0 -\loss(x)$ must be large as well.
Together \eqref{eq:bound1} explains one form of `gradient masking' \citep{papernot_practical_2016}.  Shrinking the
magnitude of gradients while also closing the gap $\loss_0 - \loss(x)$
effectively does nothing to improve adversarial robustness. 
For example, in one form of defense distillation, the magnitude of the model Jacobian is reduced
by increasing the temperature of the final softmax layer of the network.
However, this has the detrimental side-effect of sending the model output to
$(\frac{1}{N},\cdots,\frac{1}{N})$, where $N$ is the number of classes, which
effectively shrinks the loss gap to zero. An alternative form sends the
temperature of the final softmax layer to zero, which pushes the model output to
a one-hot vector; however this sends the Lipschitz constant to $\infty$. In
either form, the lower bound provided by Proposition
\ref{prop:cert} approaches zero.  

Moreover, \eqref{eq:bound2} states that even supposing the loss's gradient is small, and the gap $\loss_0-\loss(x)$ is
large, there may still be adversarially vulnerable images nearby due to errors in the
first order approximation.
Taken together, Proposition~\ref{prop:cert} provides three sufficient conditions for training robust networks:
(i) the loss gap $\ell_0 - \ell(x)$ should be large; (ii) the gradients of the loss should be small; and
(iii) the error of the first-order approximation, measured by
$\omega(\varepsilon)$, should also be small.
The first point will be satisfied by default when the loss is minimized. The
second point will be satisfied by training with a loss regularized to penalize
large input gradients.
In Section \ref{sec:exp} we provide experimental evidence that the third point
is satisfied with input gradient regularization implemented using finite
differences.

These robustness bounds are most similar in spirit to \citet{weng2018evaluating},
who derive bounds using an estimate of the \emph{local} Lipschitz constant of the
model by sampling the model gradient norm locally. Indeed, in practice, $L$ and
$\omega(\varepsilon)$ can only be estimated
and are not exactly computable on all but the simplest networks. We estimate
$L$ and $\omega(\varepsilon)$ using
the Extreme Value Theory approach outlined by \citet{weng2018evaluating}. To
estimate $L$, we sample the maximum gradient norm over batches from the data distribution, and
estimate an upper bound by fitting a generalized extreme value (GEV) distribution to
these samples through maximum likelihood estimation (MLE). Similarly, we
estimate $\omega(\varepsilon)$ by sampling the maximum of
\eqref{eq:mod} over batches of the data distribution and $\norm{v}\leq
\varepsilon$, and then fitting a GEV using MLE. The values of $L$ and
$\omega(\varepsilon)$ used in our bounds are estimated from the fitted GEV
distribution with threshold $p=0.001$.
Because these quantities are estimates, in practice the theoretical bounds of Proposition
\ref{prop:cert} can only be implemented approximately, providing a heuristic
measure of adversarial robustness.
Notably however, these bounds are valid in any norm with corresponding dual.
Furthemore, upon estimating $L$ and $\omega(\varepsilon)$,
we emphasize that these quantities only require \emph{one gradient
evaluation and one model 
evaluation} to provide an estimate on the robustness of an image.

To date, the theoretical bounds in the Euclidean norm that show the greatest
promise have arguably been those obtained through randomized smoothing
\citep{li2018second,cohen2019,salman2019}.
In randomized smoothing, models are trained with inputs perturbed through
Gaussian noise; robust prediction is achieved by averaging model predictions over
inputs again perturbed with Gaussian noise, which requires many tens of thousands of model
evaluations.
It is well known that training with
Gaussian noise is equivalent to squared $\ell_2$ norm gradient regularization
\citep{bishop1995}. Therefore we expect that models trained with squared norm gradient
regularization should have similar adversarial
robustness as those trained through randomized smoothing.


\section{Squared norm gradient regularization}\label{sec:sq}
Proposition~\ref{prop:cert} provides strong motivation for input gradient
regularization as a method for promoting adversarial robustness. However, it
does not tell us what form the gradient regularization term should take. 
In this section, we show how norm squared gradient regularization arises from a
\emph{quadratic cost}.

In adversarial training, solutions of \eqref{eq:ideal_attack} are used to
generate images on which the network is trained. In effect, adversarial training
seeks a solution of the minimax problem
\begin{eqnarray}\label{eq:robustoptim}
  \min_w \E_{x\sim\mathbb P} \left[\max_v \loss(x+v;w) - c(v) \right]
\end{eqnarray}
where $\mathbb P$ is the distribution of images.
This is a robust optimization problem \citep{wald1945statistical,
rousseeuw1987robust}.
The cost function $c(v)$ penalizes perturbed images from being too far from the
original. When the cost function is the hard constraint from
\eqref{eq:ideal_attack}, perturbations must be inside a norm ball of radius
$\delta$. This leads to
adversarial training with PGD \citep{kurakin2016adversarial,madry_2017}.
However this forces a particular scale: it is possible that no images are
adversarial within radius $\delta$, but that there are adversarial images with
only a slightly larger distance.
Instead of using a hard constraint, we can relax the cost function to be the quadratic cost $c(v) = \frac{1}{2 \delta}
\norm{v}^2$. The quadratic cost allows attacks to be of any size but penalizes
larger attacks more than smaller attacks. With a quadratic cost, there is less
of a danger that a local attack will be overlooked.

Solving \eqref{eq:robustoptim} directly is expensive: on ImageNet-1k,
both \citet{kannan2018adversarial} and   \citet{xie2018} required large-scale distributed
training with many dozens or hundreds of GPUs and over a week of training time.
Recent research suggests this hurdle may be surmounted \citep{shafahi2019,yopo}.
However, we take the view that \eqref{eq:robustoptim} may be bounded above and
solved approximately.
When the loss is smooth, the modulus of continuity is
$\omega(\varepsilon)=\varepsilon^2 C$, where $C$ is the maximum positive
 Hessian norm over all inputs.
By using the bound \eqref{eq:modulbound} with the quadratic loss
$c(v) =  \frac{1}{2 \delta}
\norm{v}^2$, the optimal value of $\max_v \loss(x+v) - c(v)$
 is
$\frac{\delta}{2(1- \delta C)} \norm{\grad_x \loss(x)}_*^2$, provided $\delta<\frac{1}{C}$.
This gives the following proposition. 
\begin{proposition}
  Suppose both the model and the loss are twice continuously differentiable.
  Suppose attacks are measured with quadratic cost $\frac{1}{2 \delta}
  \norm{v}^2$. Then the optimal value of \eqref{eq:robustoptim} is bounded above
  by
  \begin{eqnarray}\label{eq:approxbound}
    \min_w \E_{x\sim\mathbb P} \left[\loss(x;w) + \frac{\lambda}{2} \norm{\grad_x
    \loss(x)}_*^2 \right]
  \end{eqnarray}
  where $\lambda = \frac{\delta}{1 - \delta C}$.
\end{proposition}
That is, we may bound the solution of the adversarial training problem
\eqref{eq:robustoptim} by solving the gradient regularization problem
\eqref{eq:approxbound}, when the cost function is quadratic. It is not necessary
to know $\delta$ or compute $C$; they are absorbed into $\lambda$. 

Input gradient regularization using
the squared $\ell_2$ norm was proposed for adversarial robustness in \citet{ross2018}. It was expanded by \citet{roth2018} to use a
Mahalanobis norm with the correlation matrix of adversarial attacks.

When $c(v)$ is the hard constraint forcing attacks inside the $\delta$ norm
ball and $C$ is small, supposing the curvature term is negligible, we can estimate the maximum in 
\eqref{eq:robustoptim} by $\loss(x) + \frac{1}{\delta}
\norm{\grad_x \loss(x)}_*$, using  the dual norm for the gradient.  This is norm
gradient regularization (not squared) and has been used for adversarial
robustness on both CIFAR-10 \citep{gabriel2018}, and MNIST \citep{seck20191}.

\subsection{Finite difference implementation}
Norm squared input gradient regularization has long been used as a regularizer in neural
networks: \citet{drucker1991double} first showed its effectiveness for
generalization. \citeauthor{drucker1991double} implemented gradient
regularization with `double
backpropagation' to compute the derivatives of the penalty term with respect to
the model parameters $w$,
which is needed to update the parameters during training. Double backpropagation involves two passes
of automatic differentiation: one pass to compute the gradient of the loss with
respect to the inputs $x$, and another pass on the output of the first to compute the gradient of the
penalty term with respect to model parameters $w$. In neural networks, double
backpropagation is the
standard technique for computing the parameter gradient of a regularized loss.
However, it is not currently scaleable to large neural networks.
Instead we approximate the gradient regularization term with finite
differences.  
\begin{proposition}[Finite difference approximation of squared $\ell_2$ gradient norm]\label{prop:fd}
  Let $d$ be the normalized input gradient direction: 
  $d=
  \grad_x \loss(x) / \norm{\grad_x \loss(x)}_2$ when the gradient is nonzero,
  and set $d=0$ otherwise.
  Let $h$ be the finite difference step size. 
  Then, the squared $\ell_2$ gradient norm is approximated by 
\begin{eqnarray}\label{eq:fd}
    \norm{\grad_x \loss(x)}_2^2 \approx \left(\frac{\loss(x+hd) -\loss(x)}{h}\right)^2
  \end{eqnarray}
\end{proposition}
The vector $d$ is normalized to ensure the accuracy of the finite difference
approximation: as can be seen by \eqref{eq:modulbound}, the error of the finite
difference is bounded by $\omega(h) / h$.  Without normalizing the gradient, this term can
be large, which is undesirable in general. However, the finite difference error
term does play a
role in promoting adversarial robustness. Proposition
\ref{prop:cert} shows it is necessary for
$\omega(\varepsilon)$ to be small. By using finite differences, minimizing
\eqref{eq:approxbound} also implicitly minimizes $\omega(h)$, due to the error
term. Therefore the finite difference approximation
provide an indirect means for controlling the modulus of continuity
$\omega(\varepsilon)$,
necessary for adversarial robustness, through the finite difference error. This would otherwise be unavailable with
an exact method computing the gradient norm.

The finite differences approximation \eqref{eq:fd} allows the computation of the
gradient of the regularizer (with respect to model parameters $w$) to be done
with only two regular passes of backpropagation, rather than with double
backpropagation.  On the first, the input gradient direction $d$ is calculated.
The second computes the gradient with respect to model parameters by performing
backpropagation  on the right-hand-side of \eqref{eq:fd}. Double backpropagation
is avoided by detaching $d$ from the computational graph after the first pass.
In practice, for large networks, we have found that the finite difference
approximation of the regularization term is considerably more efficient than
using double backpropagation.  On networks sized for MNIST, we have observed
roughly a 10\% speed improvement by using finite differences; on CIFAR-10
networks finite differences are roughly 50\% faster than double backpropagation.

Our proposed training algorithm, with squared Euclidean input gradient
regularization, is presented in Algorithm~\ref{alg:fd} of Appendix
\ref{app:A}. Other
gradient penalty terms can be approximated as well. For example, when defending
against attacks measured in the $\ell_\infty$ norm, the squared $\ell_1$ norm
penalty can approximated by setting instead $d= \sgn(\grad_x \ell(x))/\sqrt{N}$
when the gradient is nonzero.

\begin{figure}
    \centering
    \begin{subfigure}[b]{0.48\textwidth}
      \includegraphics[height=1.6in]{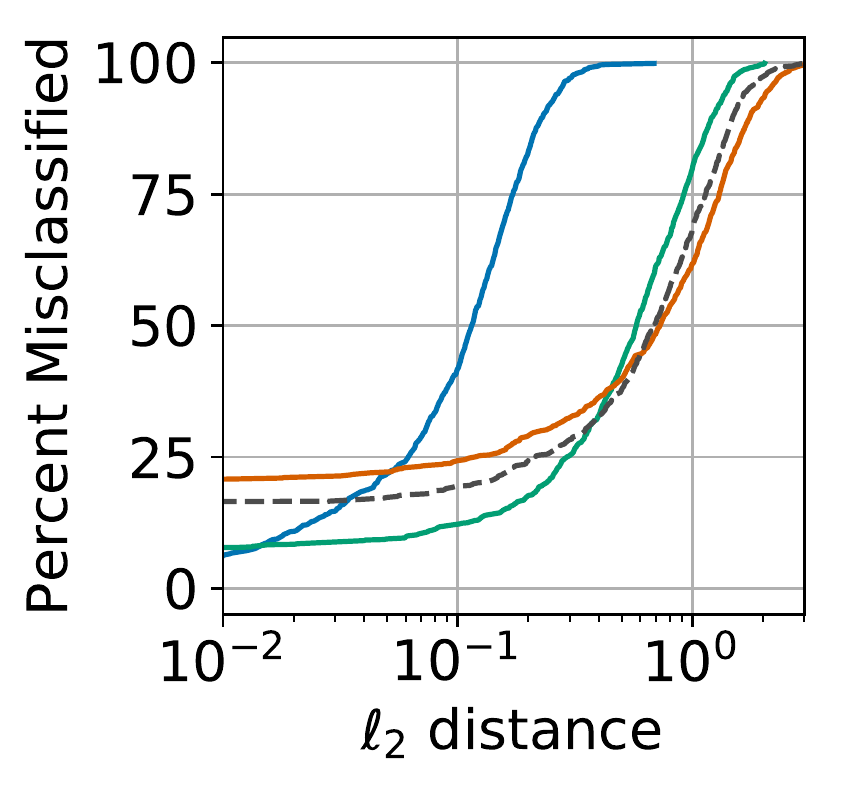}
        \caption{$\ell_2$ norm adversarial attacks}
        \label{fig:cifar10-l2}
    \end{subfigure}
    \hspace{-5em} 
    \begin{subfigure}[b]{0.48\textwidth}
      \includegraphics[height=1.6in]{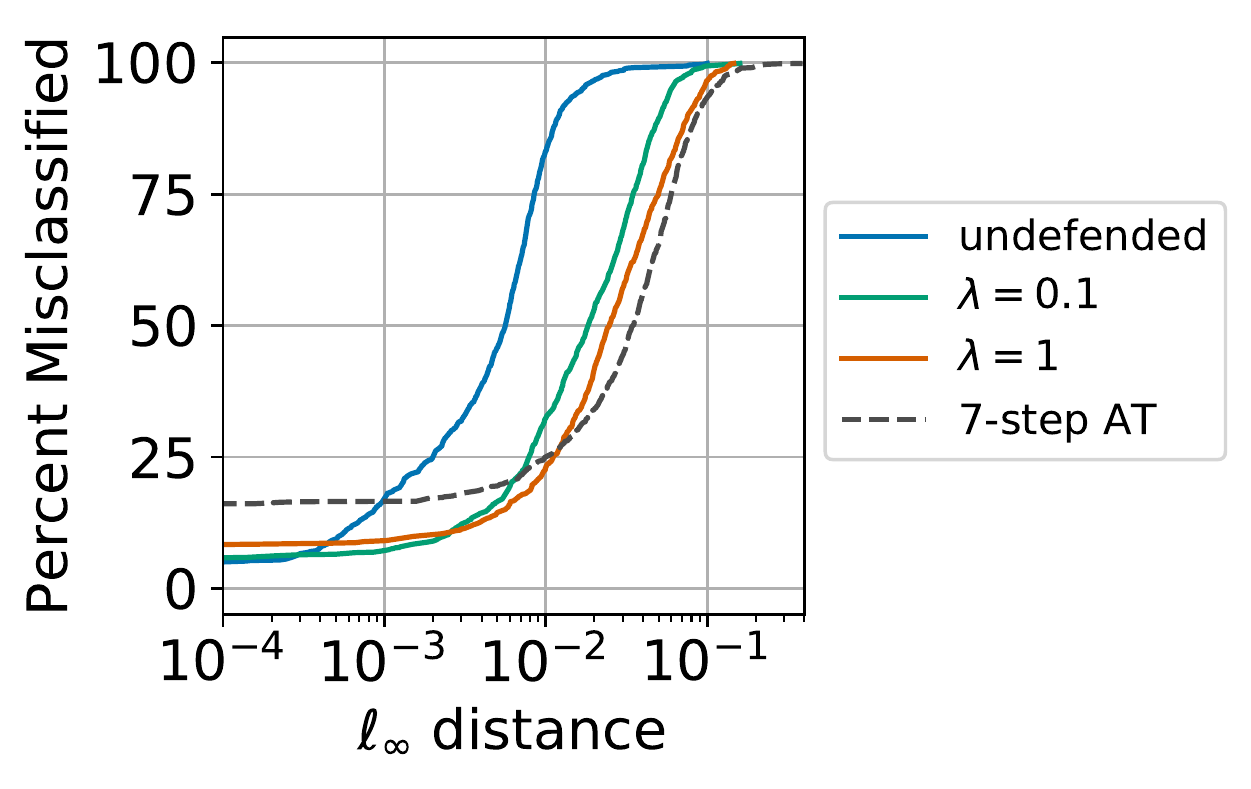}
        \caption{$\ell_\infty$ norm adversarial attacks}
        \label{fig:cifar10-linf}
    \end{subfigure}
    \caption{Adversarial attacks on the CIFAR-10 dataset, on networks built with
    standard $\relu$s. Regularized networks attacked in $\ell_2$ are trained
  with squared $\ell_2$ norm gradient regularization; networks attacked in
$\ell_\infty$ are trained with  squared $\ell_1$ norm regularization.}\label{fig:cifar10}
\end{figure} 

\section{Experimental results}\label{sec:exp}
In this section we provide empirical evidence that input
gradient regularization is an effective tool for promoting adversarial
robustness.
We attack networks with gradient-free attacks, and demonstrate gradient
regularization does not lead to gradient obfuscation. 
Moreover our results show that gradient
regularization is comparable to adversarial training, both experimentally and
through rigorous certification. 

We train networks on the CIFAR-10 dataset \citep{cifar_data}, and
ImageNet-1k \citep{imagenet}. On the CIFAR dataset we use the ResNeXt-34 (2x32)
architecture; on ImageNet-1k we use a ResNet-50 \citep{resnet}.
The CIFAR networks were trained with standard data augmentation and learning
rate schedules on a single GeForce~GTI~1080~Ti. On ImageNet-1k, we modified
the training code of \citeauthor{imagenet18}'s [\citenum{imagenet18}] submission to the DAWNBench competition
\citep{dawnbench} and train with four GPUs. Training code and trained model
weights will be made available.

We train an undefended network as a baseline to compare various types of
regularization. On CIFAR-10, networks are trained with 
squared $\ell_2$ and squared $\ell_1$ gradient norm regularization. The former is appropriate for defending against
attacks measured in $\ell_2$; the latter for attacks measured in $\ell_\infty$.
We set the regularization strength to be either $\lambda=0.1$
or 1; and set finite difference discretization $h=0.01$. 
We compare each network with the most common form of adversarial
training, where models are trained using the hyperparameters in \citet{madry_2017}
(7-steps of FGSM, projected onto an
$\ell_\infty$ ball of either radius $\frac{8}{255}$ or $\frac{2}{255}$).
On ImageNet-1k we only train adversarially robust models with squared $\ell_2$ regularization. 

On each dataset, we attack 1000 randomly selected images. We perturb each
image with attacks in both the Euclidean and $\ell_\infty$ norms,
with a suite of current state-of-the-art attacks: the Carlini-Wagner attack
\citep{carlini_towards_2016}; the Boundary attack
\citep{brendel2018decisionbased} (a gradient-free attack); the LogBarrier attack
\citep{logbarrier} (a non-local attack); and
PGD \citep{madry_2017} (in both the $\ell_\infty$ norm or the $\ell_2$ norm).
The former three attacks have all been shown effective at evading gradient
masking and gradient obfuscation.  PGD
excels at finding images close to the original when gradients are not
close to zero. We report our results using the best adversarial distance \emph{on a per image
basis}, in each norm. On the majority of test images, PGD finds the closest adversarial
image. However on a significant portion of inputs, the closest adversarial image
is found by one of the three other attacks.

\begin{table}
  \caption{Adversarial robustness statistics, measured in the $\ell_\infty$ norm.
    Top1 error is reported on CIFAR-10; Top5 error on ImageNet-1k.}
\label{tab:linf-stats}
  \begin{center}
  \begin{tabular}{lrrrrrrrc}
  \toprule
  & \multirowcell{2}{\% clean\\error}  &
  \multicolumn{2}{c}{$\omega$-bound \% error at}  & \multicolumn{2}{c}{empirical \% error at}  & \multirowcell{2}{training\\time (hours)}\\ 
  \cmidrule(lr){3-4}
  \cmidrule(lr){5-6}
                 &   &$\varepsilon=\frac{2}{255}$ & $\varepsilon=\frac{8}{255}$ &$\varepsilon=\frac{2}{255}$ & $\varepsilon=\frac{8}{255}$ &  \\
  \midrule
  \textbf{CIFAR-10}\\
  ~Undefended                                               & \textbf{4.36}  & 100 &100 &        70.82 &          98.94 &     \enspace 2.06\\
  ~$\ell_\infty$ 7-step AT, $\varepsilon=2/255$             &        6.17  & 16.02 & 100 & \textbf{16.80} &         53.90  &    10.82        \\
  ~$\ell_\infty$ 7-step AT, $\varepsilon=8/255$             &        16.33
  & 26.59 & 58.32       &         22.86 &  \textbf{46.02}
                                                                               &    12.10        \\ 
  ~squared $\ell_1$ norm, $\lambda=0.1$& 6.45 & 19.40 & 100 & 24.92 & 70.41  &    \enspace 5.22\\
  ~squared $\ell_1$ norm, $\lambda=1$  &         9.02 & 20.66 & 55.30 &        18.47 &  58.69  &    \enspace 5.15\\

  \textbf{ImageNet-1k} \\
  ~Undefended                            &  \textbf{6.94} & 100  & 100 &   90.21        &    98.94        & 20.30 \\
  ~squared $\ell_2$ norm, $\lambda=0.1$  &           7.66 & 74.7 & 100    &  70.56        &    97.53        & 32.60 \\
  ~squared $\ell_2$ norm, $\lambda=1$    &        10.26  & 63.7 & 100 & \textbf{52.79} & \textbf{95.93}  & 33.87 \\
  \bottomrule
\end{tabular}
\end{center}
\end{table}
Adversarial robustness results for networks attacked in the $\ell_\infty$ norm
are presented in Table \ref{tab:linf-stats}. These results are for networks
built of standard $\relu$s. Table \ref{tab:linf-stats} and Figure
\ref{fig:cifar10} demonstrate a clear
trade-off between test accuracy and adversarial robustness, as the strength of
the regularization is increased. On CIFAR-10, the undefended network achieves
test error of 4.36\%, but is not robust to attacks even at $\ell_\infty$
distance $\frac{2}{255}$. However with a strong regularization parameter
($\lambda=1$), test error increases to 9.02\% on clean images.
At both adversarial distance $\varepsilon=\frac{2}{255}$ and $\frac{8}{255}$
adversarial training achieves slightly better robustness (by about 5\% percent)
than our best models,
yet comes at over double the training time.

In Table \ref{tab:linf-stats} we also report heuristic measures of robustness
using \eqref{eq:bound2}. We skip reporting bounds using \eqref{eq:bound1} as
this bound returned vacuous results (an upper bound of 100\% test error at all
distances). On some test images the heuristic bound fails; this is due to the
inherent difficulty of estimating $\omega(\varepsilon)$ accurately.  



In Table \ref{tab:l2-stats} we report results on models trained for attacks
in the $\ell_2$ norm. On CIFAR-10, the most robust model is trained
with regularization strength $\lambda=1$ and outperforms even the adversarially
trained model. On ImageNet-1k, we see the same pattern: the model trained with
$\lambda=1$ offers the best protection against adversarial attacks. Due to the
long training time, we were not able to train ImageNet-1k with standard multi-step adversarial
training.
In Table \ref{tab:l2-stats} we also report our heuristic measure of  bounds on the minimum
distance required to adversarially perturb, using the Carlini-Wagner loss.\footnote{This loss can be modified
for Top-$5$ mis-classification as well.} Further $\omega$-bound results on
CIFAR-10 are presented in Table \ref{tab:cifar-wtab}.
Both \eqref{eq:bound1} and \eqref{eq:bound2} provide reasonable estimates of the
minimum adversarial distance, though \eqref{eq:bound2} is closer to
empirical results.
These heuristic measures of robustness are further corroborated by Table
\ref{tab:c2relu-reg} of Appendix \ref{app:B}, which shows that both adversarial
training and gradient regularization significantly decrease model gradients and
the modulus of continuity $\omega(\varepsilon)$.
We also use randomized smoothing, a tight and exact bound on robustness, to
verify our heuristic measures, see Table \ref{tab:cifar-cohen}.


\begin{table}
  \caption{Adversarial robustness statistics, measured in $\ell_2$.
    Top1 error is reported on CIFAR-10; Top5 error on ImageNet-1k.}
\label{tab:l2-stats}
  \begin{center}
    \begin{tabular}{lrrrrrr}
    \toprule
    & \multirowcell{2}{\% clean\\ error}   & \multicolumn{3}{c}{mean adversarial distance}&  
    \multirowcell{2}{training\\time\\(hours)} \\[1ex]
    \cmidrule(lr){3-5}
     &&$L$-bound & $\omega$-bound & empirical & & \\
  \midrule 
  \textbf{CIFAR-10} \\
  ~Undefended                            & \textbf{4.36} &0.006  & 0    &    0.12         & \enspace2.06 \\
  ~$\ell_\infty$ 7-step AT, $\varepsilon=2/255$               & 6.17 &0.05            & 0.36  &    0.66         & 10.82        \\
  ~$\ell_\infty$ 7-step AT, $\varepsilon=8/255$               & 16.33         &0.18            & \textbf{0.55}  &    0.74         &12.10        \\
  ~squared $\ell_2$ norm, $\lambda=0.1$    &  8.03         &0.14          & 0.34  &    0.63         &\enspace5.18 \\
  ~squared $\ell_2$ norm, $\lambda=1$      & 20.31         &\textbf{0.30} & 0.50  &    \textbf{0.81}& \enspace5.08 \\
  \textbf{ImageNet-1k} \\                                                                         
  ~Undefended                 &\textbf{6.94}  &\num{3.63e-02}&           0.5  &   0.55&20.30 \\ 
  ~squared $\ell_2$ norm,  $\lambda=0.1$   & 7.66  &0.13& 0.96 &    1.14 &32.60 \\ 
  ~squared $\ell_2$ norm,  $\lambda=1$     &10.26 &\textbf{0.26}& \textbf{1.23} &    \textbf{1.75}&33.87 \\ 
  \bottomrule

\end{tabular}
\end{center}
\end{table}

\section{Conclusion}
We have provided motivation for training adversarially robust networks through
input gradient regularization, by bounding the minimum adversarial distance with
gradient statistics of the loss. We have shown that gradient regularization is
scaleable to ImageNet-1k and provides adversarial robustness competitive with
adversarial training, both theoretically and empirically.
Moreover, we have demonstrated that gradient regularization does not lead to
gradient obfuscation.
Networks may be trained using gradient regularization in fractions of the
training time of standard adversarial training.

We gave theoretical per-image bounds on the minimum adversarial distance for
non-smooth models using the Lipschitz constant of the loss and its modulus of
continuity, which measures the error of a first order approximation of the loss.
These bounds were empirically compared against state-of-the-art
attacks and other theoretical bounds. 
Although in practice these bounds give heuristic measures of robustness, they are valid in
any norm and provide per-image robustness measures at the cost of only one model
evaluation and one gradient evaluation.

{\small
\bibliographystyle{iclr2020_conference}
\bibliography{DNN}
}

%
\newpage
\appendix
\section{Training algorithm}\label{app:A}

\begin{algorithm}[H]
  \caption{Training with squared $\ell_2$-norm input gradient regularization, using finite differences}
   \label{alg:fd}
   \begin{algorithmic}[1]
  \STATE {\bfseries Input:} Initial model parameters $w_0$ \\
{\bfseries Hyperparameters:} Regularization strength $\lambda$; batch size $m$;
  finite difference discretization $h$
\WHILE{$w_t$ not converged}
\STATE sample minibatch of data $\left\{ (x^{(i)}, y^{(i)}) \right\}_{i=1,\dots,m}$
  from empirical distribution $\hat{\mathbb P}$
  \FOR{$i=0$ {\bfseries to} $m$}
    \STATE  $g^{(i)} =  \nabla_x \loss(x^{(i)},y^{(i)}; w_t)$
    \STATE \(  d^{(i)} =
           \begin{cases}
             \frac{g^{(i)}}{\norm{g^{(i)}}_2} & \text{ if } g^{(i)} \neq 0 \\
             0 & \text{ otherwise}
           \end{cases}
         \)
         \COMMENT{for $\ell_1$-norm use normalized signed gradient}
    \STATE detach $d^{(i)}$ from computational graph
    \STATE $z^{(i)} = x^{(i)} + h d^{(i)}$
  \ENDFOR
  \STATE  $\mathcal L(w) = \frac{1}{m} \sum_{i=1}^m \loss(x^{(i)}, y^{(i)}; w)$
  \vspace{0.5em}
  \STATE  $\mathcal R(w) = \frac{1}{m} \sum_{i=1}^m \frac{1}{h^2} \left(\loss(z^{(i)}, y^{(i)}; w) -
  \loss(x^{(i)}, y^{(i)}; w)\right)^2$
  \vspace{0.5em}
  \STATE $w_{t+1} \leftarrow w_t - \tau_t \grad_w\left( \mathcal L(w_t) + \lambda
  \mathcal R(w_t)\right )$ 
   \ENDWHILE
\end{algorithmic}
\end{algorithm}

\section{Additional empirical results}\label{app:B}

\begin{figure}[H]
    \centering
    \begin{subfigure}[b]{0.48\textwidth}
      \includegraphics[height=1.5in]{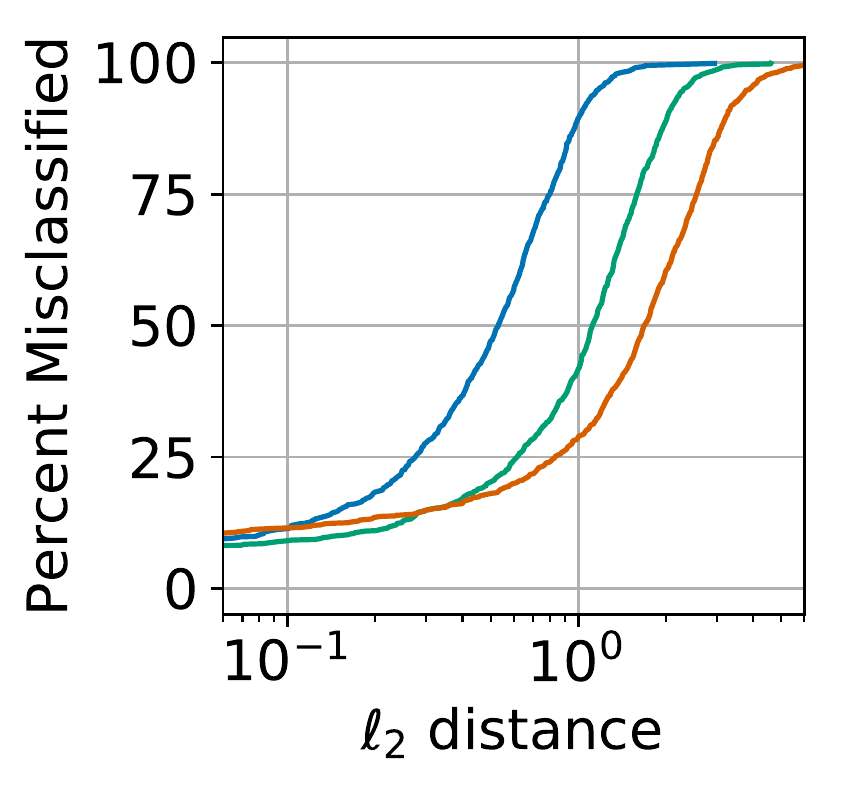}
        \caption{$\ell_2$-norm adversarial attacks}
        \label{fig:imagenet-l2}
    \end{subfigure}
    \hspace{-5em}
    ~ 
    \begin{subfigure}[b]{0.48\textwidth}
      \includegraphics[height=1.5in]{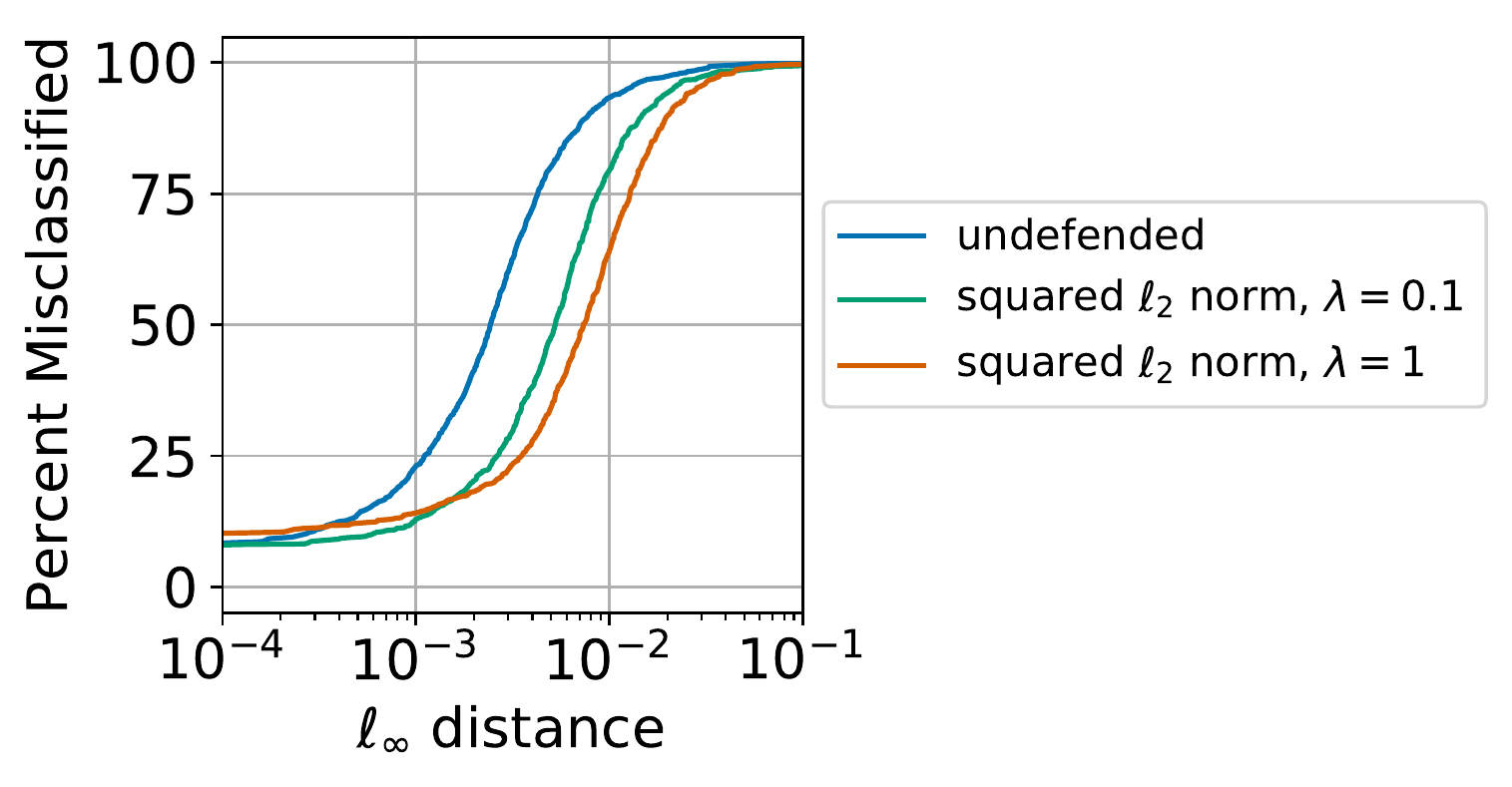}
        \caption{$\ell_\infty$-norm adversarial attacks}
        \label{fig:imagenet-linf}
    \end{subfigure}
    \caption{Adversarial attacks on ImageNet-1k with the ResNet-50 architecture. Top5 error reported.}\label{fig:imagenet}
\end{figure}

%

\begin{table}
  \caption{Regularity statistics on selected models, measured in the $\ell_2$
    norm. Statistics computed using modified loss $\max_{i\neq c} f_i(x)- f_c(x)$ . A
 soft maximum is used for curvature statistics.}
  \begin{center}
    \begin{tabular}{lrrrr}
    \toprule
    &   \multicolumn{2}{c}{$\norm{\grad \loss(x)}$}  &\multirowcell{2}{$\omega(0.5)$}&\multirowcell{2}{$\omega(1.0)$} \\ 
    \cmidrule(lr){2-3} 
    &mean & max &   \\
  \midrule
  \textbf{CIFAR-10}\\
  ~Undefended                                      & 3.05  &  122.34 & 1.80 & 3.13  \\
  ~$\ell_\infty$ 7-step AT, $\varepsilon=2/255$    & 0.63  &  11.96  & 0.08 & 0.32  \\
  ~$\ell_\infty$ 7-step AT, $\varepsilon=8/255$    & 0.40  &  2.52   & 0.02 & 0.06  \\
  ~squared $\ell_2$ norm, $\lambda=0.1$            & 0.58  &  4.43   & 0.08 & 0.50  \\
  ~squared $\ell_2$ norm, $\lambda=1$              & 0.35  &  1.33   & 0.02 & 0.05  \\
  \textbf{ImageNet-1k}\\
  ~Undefended                                      & 1.12 & 17.51   & \num{1.7e-2} & \num{6.7e-2}       \\
  ~squared $\ell_2$ norm, $\lambda=0.1$            & 0.46 & 4.85    & \num{5.5e-3} & \num{2.2e-2}      \\
  ~squared $\ell_2$ norm, $\lambda=1$              & 0.27 & 2.12    & \num{3.4e-3} & \num{2.1e-2}      \\
  \bottomrule

\end{tabular}
\end{center}
\label{tab:c2relu-reg}
\end{table}

\begin{table}
  \caption{Certified test error (\%) at various $\ell_2$
    radii on CIFAR-10, using the randomized smoothing certification technique of \citet{cohen2019}.}
  \begin{center}
    \begin{tabular}{lrrrrrrrr}
    \toprule
    & 0.25 & 0.5 & 0.75 & 1.0 & 1.25 & 1.5 & 1.75 & 2.0 \\ 
  \midrule
  Undefended & 90 & 92 & 96 & 99 & 100 & 100 & 100 & 100 \\
  squared $\ell_2$ norm, $\lambda=1$ & 47 & 63 & 80 & 90 & 90 & 90& 90 & 90 \\ 
  \midrule 
  \citet{cohen2019} & 40 & 57 & 68 & 77 & 83 & 86 & 88 & 90 \\
  \citet{salman2019} &  26 & 43 & 52 & 62 & 67 & 71 & 75 & 81 \\
  \bottomrule

\end{tabular}
\end{center}
\label{tab:cifar-cohen}
\end{table}

\begin{table}
  \caption{Certified test error (\%) at various $\ell_2$
  radii on CIFAR-10, using \eqref{eq:bound2}.}
  \begin{center}
    \begin{tabular}{lrrrrrrrr}
    \toprule
    & 0.25 & 0.5 & 0.75 & 1.0 & 1.25 & 1.5  \\ 
  \midrule
  Undefended & 100 & 100 & 100 & 100 & 100 & 100  \\
$\ell_\infty$ 7-step AT, $\varepsilon=2/255$ & 19 & 91 & 100 & 100 & 100&100\\
$\ell_\infty$ 7-step AT, $\varepsilon=8/255$ & 31 & 47 & 61 & 77 & 100 & 100 \\
  squared $\ell_2$ norm, $\lambda=0.1$ & 22 & 99 & 100 & 100& 100 & 100  \\ 
  squared $\ell_2$ norm, $\lambda=1$ & 33 & 50 & 65 & 80 & 98 & 100  \\ 
  \bottomrule

\end{tabular}
\end{center}
\label{tab:cifar-wtab}
\end{table}

%

\end{document}